\documentclass{article}
\usepackage[width=122mm,left=12mm,paperwidth=146mm,height=193mm,top=12mm,paperheight=217mm]{geometry}

\usepackage[utf8]{inputenc} % allow utf-8 input
\usepackage[T1]{fontenc}    % use 8-bit T1 fonts
\usepackage{hyperref}       % hyperlinks
\usepackage{url}            % simple URL typesetting
\usepackage{booktabs}       % professional-quality Tables
\usepackage{amsfonts}       % blackboard math symbols
\usepackage{nicefrac}       % compact symbols for 1/2, etc.
\usepackage{microtype}      % microtypography
\date{}
\usepackage{times}
\usepackage{helvet}
\usepackage{courier}
\usepackage{amsmath}
\usepackage{amssymb}
\usepackage{amsthm}
\usepackage{graphicx}
\usepackage{array}
\usepackage{booktabs}
\usepackage{amsopn}
\usepackage{amsmath,bm}
\usepackage{algorithm}
\usepackage{algorithmic}
\usepackage{enumerate}
\usepackage{enumitem}
\usepackage{color}
\usepackage{multirow}

\floatstyle{plain}
\newfloat{myalgo}{tbhp}{mya}
\newenvironment{Algorithm}[1][tbh]%
{\begin{myalgo}[#1]
\centering
\begin{minipage}{11.5cm}
\begin{algorithm}[H]}%
{\end{algorithm}
\end{minipage}
\end{myalgo}}

\newcommand{\eg}{\emph{e.g. }}
\newcommand{\etal}{\emph{et al. }}
\newcommand{\ie}{\emph{i.e. }}

\newcommand{\wrt}{w.r.t. }

\graphicspath{{figs/}}

\title{Towards Evolutional Compression}

\author{\small{Yunhe Wang$^{1}$, Chang Xu$^{2}$, Jiayan Qiu$^{2}$, Chao Xu$^{1}$, Dacheng Tao$^{2}$}\\
  \scriptsize{$^{1}$Key Laboratory of Machine Perception (MOE), Cooperative Medianet Innovation Center, School of EECS, Peking University}\\
   \scriptsize{ $^{2}$ UBTech Sydney AI Centre, School of IT, FEIT, The University of Sydney, Australia}\\
  \texttt{\scriptsize{wangyunhe@pku.edu.cn, c.xu@sydney.edu.au, jqiu3225@uni.sydney.edu.au,}}\\
  \texttt{\scriptsize{xuchao@cis.pku.edu.cn, dacheng.tao@sydney.edu.au, }} \\}
  
\begin{document}

\maketitle

\begin{abstract}
Compressing convolutional neural networks (CNNs) is essential for transferring the success of CNNs to a wide variety of applications to mobile devices. In contrast to directly recognizing subtle weights or filters as redundant in a given CNN, this paper presents an evolutionary method to automatically eliminate redundant convolution filters. We represent each compressed network as a binary \emph{individual} of specific fitness. Then, the \emph{population} is upgraded at each evolutionary iteration using genetic operations. As a result, an extremely compact CNN is generated using the fittest individual. In this approach, either large or small convolution filters can be redundant, and filters in the compressed network are more distinct. In addition, since the number of filters in each convolutional layer is reduced, the number of filter channels and the size of feature maps are also decreased, naturally improving both the compression and speed-up ratios. Experiments on benchmark deep CNN models suggest the superiority of the proposed algorithm over the state-of-the-art compression methods.
\end{abstract}

\section{Introduction}
Large-scale deep convolutional neural networks (CNNs) have been successfully applied to a wide variety of applications such as image classification~\cite{ResNet,VGGNet, AlexNet}, object detection~\cite{RCNN,fasterRCNN}, and visual enhancement~\cite{SRCNN}. To strengthen representation capability and improve CNN performance, several convolutional layers have traditionally been used in network construction. Given the complex network architecture and numerous variables, most CNNs place excessive demands on storage and computational resources, thus limiting them to high-performance servers.

We are now in an era of intelligent mobile devices. Deep learning, one of the most promising artificial intelligence techniques, is expected to reduce reliance on servers and to apply advanced algorithms to smartphones, tablets, and wearable computers. Nevertheless, it remains challenging for mobile devices without GPUs and the necessary memory to carry CNNs usually run on servers. For instance, more than 232\emph{MB} of memory and $7.24\times10^8$ floating number multiplications would be consumed by AlexNet~\cite{AlexNet} or VGGNet~\cite{VGGNet} to process a single, normal-sized input image. Hence, special developments are required to translate CNNs to smartphones and other portable devices.

To overcome this conflict between reduced hardware configurations and the higher resource demands of CNNs, several attempts have been made to compress and speed up well-trained CNN models. Liu \etal developed to~\cite{sparseNet} learn CNNs with sparse architectures, thereby reducing model complexities compared to ordinary CNNs, while Han \etal~\cite{pruning15} directly discarded subtle weights in pre-trained CNNs to obtain sparse CNNs. Figurnov \etal~\cite{sparse1} reduced the computational cost of CNNs by masking the input data of convolutional layers. Wen \etal~\cite{sparse2} explored subtle connections from the perspective of channels and filters. In the DCT frequency domain, Wang \etal~\cite{CNNpackNIPS} excavated redundancy on all weights and their underlying connections to deliver higher compression and speed-up ratios. In addition, a number of techniques exist to compress convolution filters in CNNs including weight pruning~\cite{pruning15,pruning}, quantization and binarization~\cite{quan2,binary2,3binary,binary}, and matrix decomposition~\cite{SVD}.

While these methods have reduced the storage and computational burdens of CNNs, the research to deep model compression is still in its infancy. Existing solutions are typically grounded in different, albeit intuitive, assumptions of network redundancy, \eg weight or filter redundancy with small absolute values, low-rank filter redundancy, and within-weight redundancy. Although these redundancy assumptions are valid, we hypothesize that all possible types of redundancy have yet to be identified and validated. We postulate that an approach can be developed to autonomously investigate the diverse and volatile network redundancies and to constantly upgrade the solution to cater for environment changes.

In this paper, we develop an evolutionary strategy to excavate and eliminate redundancy in deep neural networks. The network compression task can be formulated as a binary programming problem, where a binary variable is attached to each convolution filter to indicate whether or not the filter takes effect. Inspired by studies in evolutionary computation~\cite{GA,GAgoogle,SA}, we treat each binary encoding \wrt a compressed network as an individual and stack them to constitute the population. A series of evolutionary operators (\eg crossover and mutation) allow the population to constantly evolve to reach the most competitive, compact, and accurate network architecture. When evaluating an individual, we use a relatively small subset of the original training dataset to fine-tune the compressed network, which quickly excavates its potential performance; therefore, the overall running times of compressing CNNs are acceptable. Experiments conducted on several benchmark CNNs demonstrate that compressed networks are more lightweight but have comparable accuracies to their original counterparts. Beyond conventional network redundancies, we suggest that convolutional filters with either large or small weights possess redundancies, the discrepancy between filters is appreciated, and that high-frequency coefficients of convolution filters are unnecessary (\ie smooth filters are adequate). 

\begin{figure}[t]
\centering
%\rule{12cm}{8cm}
\includegraphics[width=0.99\linewidth]{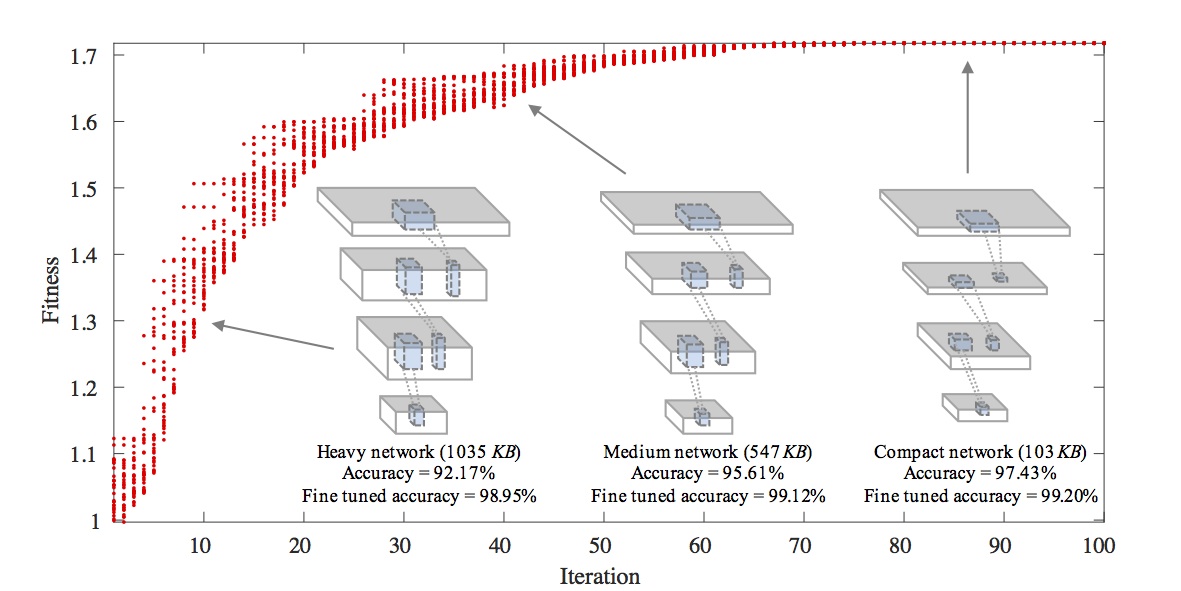}
\vspace{-0.5em}
\caption{An illustration of the evolution of LeNet on the MNIST dataset. Each dot represents an \emph{individual} in the population, and the thirty best individuals are shown in each evolutional iteration. The fitness of individuals is gradually improved with an increasing number of iterations, implying that the network is more compact but remaining the same accuracy. The size of the original network is about 1.5\emph{MB}.}
\label{Fig:diagram}
\vspace{-0.5em}
\end{figure}

\section{An Evolutionary Method for Compressing CNNs}
Most existing CNN compression methods are based on the consensus that weights or filters with subtle values have limited influence on the performance of the original network. In this section, we introduce an evolutionary algorithm to significantly excavate redundancy in CNNs and devise a novel compression method.
 
\subsection{Molding Redundancy in Convolution Filters}
Considering a convolutional neural network $\mathcal{L}$ with $p$ convolutional layers $\{\mathcal{L}_1,...,\mathcal{L}_p \}$, we define $p$  sets of convolution filters $\mathbf{F} = \{F_1,...,F_p \}$ for these layers. For the $i$-th convolutional layer, its filter is denoted as $F_i\in\mathbb{R}^{H_i\times W_i\times C_i\times N_i}$, where $H_i$ and $W_i$ are the height and width of filters, respectively, $C_i$ is the channel size, and $N_i$ is the number of filters in this layer. Given a training sample $\mathcal{X}$ and the corresponding ground truth $\mathcal{Y}$, the error of the network can be defined as $E(\mathcal{X},\mathcal{Y},\mathbf{F})$, which could be, for example, softmax or Euclidean losses. The conventional weight pruning algorithm can be formulated as 
\begin{equation}
\begin{aligned}
\min_{B_1,...,B_p}\; ||E(\mathcal{X},\mathcal{Y},\hat{\mathbf{F}})-\mathcal{Y}||_F^2&+\lambda\sum_{i=1}^p||B_i||_1,\\
s.t. \quad \hat{\mathbf{F}} = \{F_1\circ B_1,...,F_p\circ B_p\}, \; B_i\in&\{0,1\}^{H_i\times W_i\times C_i\times N_i},\; \forall\; i = 1,...,p,
\end{aligned}
\label{Fcn:obj1}
\end{equation}
where $B_i$ is a binary tensor for removing redundant weights in $\mathbf{F}$, $||\cdot||_1$ is the $\ell_1$-norm accumulating absolute values $B_i$, \ie the number in $B_i$, $\circ$ is the element-wise product, and $\lambda$ is the tradeoff parameter. A larger $\lambda$ will make $B_i$ more sparse and so a network parameterized with $\mathbf{F}$ will have fewer weights.

In general, Fcn.~\ref{Fcn:obj1} is easy to solve if $E(\mathcal{X},\mathcal{Y},\hat{\mathbf{F}})$ is a linear mapping of $\hat{\mathbf{F}}$. However, neural networks are composed of a series of complex operations, such as pooling and ReLU, which increase the complexity of Fcn.~\ref{Fcn:obj1}. Therefore, a greedy strategy~\cite{pruning,sparse2} have been introduced to obtain a feasible solution that removes weights with small absolute values: 
\begin{equation}
B_i^{(h,w,c,n)} = \bigg\{
\begin{array}{l}
0, \quad \text{if}\ \ |F_i^{(h,w,c,n)}| \leq \tau,\\
1, \quad \text{otherwise}, 
\end{array}
\label{Fcn:B1}
\end{equation}
where $\tau>0$ is a threshold. This strategy is based on the intuitive assumption that small weights make subtle contributions to the calculation of the convolution response.

Although sparse filters learned by Fcn.~\ref{Fcn:obj1} demand less storage and computational resources, the size of the feature maps produced by these filters does not change. For example, a convolutional layer with 10 filters will generate 10 feature maps for one input data before and after compression, which accounts for a large proportion of online memory usage. Moreover, these sparse filters usually need some additional techniques to support and speed-up their compression such as CuSparse kernel, CSR format, or the fixed-point multiplier~\cite{pruning}. Therefore, more flexible approaches~\cite{trimming,sparse2,sparse1} have been developed to directly discard redundant filters:
\begin{equation}
B_i^{(:,:,:,n)} = \bigg\{
\begin{array}{l}
0, \quad \text{if}\ \ ||F_i^{(:,:,:,n)}||_F^2 \leq \tau,\\
1, \quad \text{otherwise},
\end{array}
\label{Fcn:B2}
\end{equation}
where $B_i^{(:,:,:,n)}$ denotes the $n$-th filter in the $i$-th convolutional layer. By directly removing convolution filters, network complexity can be significantly decreased. However, Fcn.~\ref{Fcn:B2} is also biased since the Frobenius norm of filters is not a reasonable redundancy indicator. For example, most of the weights in a filter for extracting edge information are very small. Thus, a more accurate approach for identifying redundancy in CNNs is urgently required.

\subsection{Modeling Redundancy by Exploiting Evolutionary Algorithms}
Instead of the greedy strategies shown in Fcns.~\ref{Fcn:B1} and~\ref{Fcn:B2}, evolutionary algorithms such as the genetic algorithm (GA~\cite{GA,GAgoogle}) and simulated annealing (SA~\cite{SA}) have been widely applied to the NP-hard binary programming problem. A series of bit (0 or 1) strings (\emph{individuals}) are used to represent possible solutions of the binary programming problem, and these individuals evolve using some pre-designed operations to maximize their \emph{fitnesses}.

A binary variable can be attached to each weight in the CNN $\mathcal{L}$ to indicate whether the weight takes effect or not, but a large number of binary variables will significantly slow down the CNN compression process, especially for sophisticated CNNs learned over large-scale datasets (\eg ImageNet~\cite{ImageNet}). For instance, AlexNet~\cite{AlexNet} has eight convolutional layers with more than $6\times10^7$ 32-floating weights in total, so it is infeasible to generate a population with hundreds of $6\times10^7$-dimensional individuals. In addition, as mentioned above, excavating redundancy in convolution filters itself produces a regular CNN model with less computational complexity and memory usage, which is more suitable for practical applications. Therefore, we propose to assign a binary bit to each convolution filter in a CNN, and these binary bits form an \emph{individual} $\mathbf{b}$ in this network. By doing so, the dimensionality of $\mathbf{b}$ is tolerable, \eg $\mathbf{b}\in\{0,1\}^{9568}$ (without the last 1000 convolution filters corresponding to the 1000 classes in the ILSVRC 2012 dataset) for AlexNet.

During evolution, we use GA to constantly update individuals of greater fitness. Other evolutionary algorithms can be applied using a similar approach. The compression task has two objectives: preserving performance and removing the redundancy of the original networks. The fitness of a specific \emph{individual} $\mathbf{b}$ can therefore be defined as 
\begin{equation}
f(\mathbf{b}) =  1-E(\mathcal{X},\mathcal{Y},\hat{\mathbf{F}})+\frac{\lambda}{L}\sum_{i=1}^p||1-\mathbf{b}_i||_1,
\label{Fcn:fit1}
\end{equation}
where $\mathbf{b}_i$ denotes the binary bit for the $i$-th convolution filter in the given network, and $L$ is the number of all convolution filters in the network. $E(\mathcal{X},\mathcal{Y},\hat{\mathbf{F}})$ calculates the classification loss of the network using compressed filters $\hat{\mathbf{F}} = \{F_1\circ B_1,...,F_p\circ B_p\}$, which supposed as a general loss taken value from 0 to 1. In addition, we include a constant $1$ in Fcn.~\ref{Fcn:fit1}, which ensures $f(\mathbf{b})>0$ for the convenience of calculating the probability of each individual in the evolutionary algorithm process. $\lambda>0$ is the tradeoff parameter, and
\begin{equation}
B_i^{(:,:,:,n)} = \bigg\{
\begin{array}{l}
0, \quad \text{if}\ \ \mathbf{b}_i(n)=0,\\
1, \quad \text{otherwise},
\end{array}
\end{equation}
where $\mathbf{b}_i(n) = 0$ implies that the $n$-th filter in the $i$-th layer has been discarded, otherwise retained.

\begin{Algorithm}[t]
\caption{ECS: Evolution method for compressing CNNs.}
\label{Alg:main}
%\small
\begin{algorithmic}[1]
\REQUIRE An image dataset $\{\mathcal{X},\mathcal{Y}\}$ including $n$ images for evaluating individuals, a pre-trained network $\mathcal{L}$, parameters: the scale of the population $K$, the maximum iteration number $T$, $\lambda$, $s_1$, $s_2$, and $s_3 = 1-s_1-s_2$.
\STATE Randomly initialize the population $P_1$, each individual is represented as a binary vector \wrt convolution filters in the given network $\mathcal{L}$;
\FOR{$t = 2$ to $T$}
\STATE Calculate the fitness of each individual in $P_{t-1}$ using Fcn.~\ref{Fcn:fit3};
\STATE Calculate probability of being selected of individuals according to Fcn.~\ref{Fcn:prob};
\FOR{$k = 2$ to $K$}
\STATE $P_{t}^{(1)}\leftarrow$ the best individual in $P_{t-1}$;
\STATE Generate a random value which follows a uniform distribution $s\sim[0,1]$;
\IF{$s<s_1$}
\STATE $P_{t}^{(k)}\leftarrow$ a randomly selected parent according to Fcn.~\ref{Fcn:prob};
\ELSIF {$s1\leq s<s_1+s_2$}
\STATE Randomly select two parents and generate two offspring;
\STATE Calculate fitnesses of offspring, $P_{t}^{(k)}\leftarrow$ the best offspring;
\ELSE 
\STATE $P_{t}^{(k)}\leftarrow$ a randomly selected parent after applying XOR on a fragment;
\ENDIF
\ENDFOR
\ENDFOR
\STATE Use the optimal individual in $P_T$ to construct a compact neural network $\hat{\mathcal{L}}$;
\ENSURE The compressed $\hat{\mathcal{L}}$ after fine-tuning.
\end{algorithmic}
\end{Algorithm}

In addition, the last term in Fcn.~\ref{Fcn:fit1} implicitly assumes that discarding every convolution filter makes an equivalent contribution to compression. However, the memory utilization of filters in different convolutional layers is different and related to the height $H$, width $W$, and the number of channel $C$. Therefore, filter size must be taken into account, and Fcn.~\ref{Fcn:fit1} can be reformulated as: 
\begin{equation}
f(\mathbf{b}) =  1-E(\mathcal{X},\mathcal{Y},\hat{\mathbf{F}})+\frac{\lambda}{M}\sum_{i=1}^p\left( H_iW_iC_i\cdot||1-\mathbf{b}_i||_1\right),
\label{Fcn:fit2}
\end{equation}
where $H_i$, $W_i$, $C_i$ are height, width, and channel number of filters in the $i$-th convolutional layer, respectively. $M = \sum_{i=1}^p H_iW_iC_iN_i$ is the total number of weights in the network, which scales the last term in Fcn.~\ref{Fcn:fit2} to $[0,\lambda]$.

In addition, the number of channels $C_i$ in the $i$-th layer is usually set as the number of convolution filters $N_{i-1}$ ($N_0 = 3$ for RGB color images) in the $(i-1)$-th layer to make two consecutive network layers compatible. Instead of fixing $C_i$ in Fcn.~\ref{Fcn:fit2}, $C_i$ should various with $\mathbf{b}_{i-1}$. Thus, we reformulate the calculation of fitness as follows: 
\begin{equation}
f(\mathbf{b}) =  1-E(\mathcal{X},\mathcal{Y},\hat{\mathbf{F}})+\frac{\lambda}{M}\sum_{i=1}^p\left( H_iW_i\cdot||1-\mathbf{b}_{i-1}||_1\cdot||1-\mathbf{b}_i||_1\right),
\label{Fcn:fit3}
\end{equation}
where $\mathbf{b}_p(n) = 1$, $\forall$ $n = 1,...,N_p$ for the last layer consisting of nodes corresponding to different classes in a particular dataset. The second objective in Fcn.~\ref{Fcn:fit3} accumulates the discarded weights of the compressed network. Since the error rate of a network tends to be influenced by adjusting the network architecture, we use a subset of the training data (10k images randomly extracted from the training set) to fine-tune the network weights and then re-calculate $E(\mathcal{X},\mathcal{Y},\hat{\mathbf{F}})$ to provide a more reasonable evaluation. This fine-tuning is fast, since compressed networks with fewer filters require much less computation, \eg fine-tuning over 10k images will cost about 2 seconds for LeNet and about 30 seconds for AlexNet, which is tolerable. Then, GA is deployed to discover the fittest individual through several evolutions detailed in the next section.

\subsection{Genetic Algorithm for Optimization}
GA can automatically search for compact neural networks by alternately evaluating the fitness of each individual in the whole population and executing operations on individuals. The population in the current iteration are regarded as parents, who breed another population as offspring using some representative operations, including \textbf{selection}, \textbf{crossover}, and \textbf{mutation}, with the expectation that the subsequent offsprings are fitter than the preceding parents. First, each individual is given a probability by comparing its fitness against those of other \emph{individuals} in the current \emph{population}:  
\begin{equation}
\Pr(\mathbf{b}^j) =  \left. f(\mathbf{b}^j) \middle/ \sum_{k=1}^{K}f(\mathbf{b}^k) \right.,
\label{Fcn:prob}
\end{equation}
where $K$ is the number of \emph{individuals} in the \emph{population}. Then, the above three operations will be randomly applied as follows:

\textbf{Selection.}  Given a probability parameter $s_1$, an individual is selected according to Fcn. ~\ref{Fcn:prob} and then directly duplicated as an offspring. It is clear that compressed networks with higher accuracy and compression ratios will be preserved. The best individual in the parent population is usually inherited to preserve the optimal solution.

\textbf{Crossover.} Given a probability parameter $s_2$, two selected parents according to Fcn.~\ref{Fcn:prob} will be crossed to generate two offspring as follows: 
\begin{equation*}
\footnotesize
\begin{aligned}
\text{parent}_1:  1011101010\Big|\mathbf{0101110010}\Big| 010100&\quad\;\;\;\; \text{parent}_2:  1010001011\Big|\mathbf{1010101011}\Big| 011010\\
\text{offspring}_1:  1011101010\Big|\mathbf{1010101011}\Big| 010100&\quad \text{offspring}_2:  1010001011\Big|\mathbf{0101110010}\Big| 011010
\end{aligned}
\end{equation*}
The objective of the crossover operation is to integrate excellent information from the parents. The fitness of two offspring are different, and we discard the weaker one.

\textbf{Mutation.} To promote population diversity, mutation randomly changes a fragment in the parent and produces an offspring. The conventional mutation operation for binary encoding is a XOR operation as follows: 
\begin{equation*}
\footnotesize
\begin{aligned}
\text{parent}:  0110100\Big|\mathbf{1001010100001}\Big| 1010100\quad\quad\text{offspring}:  0110100\Big|\mathbf{0110101011110}\Big| 1010100
\end{aligned}
\end{equation*}
The parent is also selected according to Fcn.~\ref{Fcn:prob}, and the mutation operation is performed with a probability parameter . Since the scale of offspring is the same as that of parents, we have $s_1+s_2+s_3 = 1$.

By iteratively employing these three genetic operations, the initial population will be updated efficiently until the maximum iteration number is achieved. After obtaining the individual with optimal fitness, we can reconstruct a compact CNN. Then, the fine-tuning strategy is adopted to enhance the performance of the compressed network. Alg.~\ref{Alg:main} summarizes the proposed evolutionary method for compression.

\section{Analysis of Compression and Speed-up Improvements}
In the above section, we presented the evolutionary method for compressing pre-trained CNN models. Since there is at least one convolution filter in the compressed network $\hat{\mathcal{L}}$, it has the same depth but less filters in $\hat{\mathbf{F}}$ compared to the original network $\mathcal{L}$ with $\mathbf{F}$. Here we further analyze the memory usage and computational cost of compressed CNNs using Alg.~\ref{Alg:main}.

\textbf{Speed-up ratio.} For a given image, the $i$-th convolutional layer $\mathcal{L}_i$ produces feature maps $Y_i\in\mathbb{R}^{H'_i\times W'_i \times N_i}$ through a set of convolution filters $F\in\mathbb{R}^{H_i\times W_i\times C_i\times N_i}$, where $H'_i$ and $W'_i$ are the height and width of feature maps, respectively, and $C_i = N_{i-1}$. Since multiplications of 32-bit floating values are much more expensive than additions, and there is more computation in other auxiliary layers (\eg pooling, ReLU, and batch normalization), speed-up ratios are usually calculated on these floating number multiplications~\cite{XNORNet,CNNpackNIPS}. Considering the major computational cost, the speed-up ratio of the compressed network for this layer compared to the original network is
\begin{equation}
r_{s_i} = \frac{H_i W_i N_{i-1}N_iH'_iW'_i}{H_iW_i\hat{N}_{i-1}\hat{N}_iH'_iW'_i} =  \frac{N_{i-1}N_i}{\hat{N}_{i-1}\hat{N}_i},
\label{Fcn:rs}
\end{equation}
where $\hat{N}_i = ||1-\mathbf{b}_i||_1$ is the number of filters in the $i$-th convolutional layer of the compressed network, as shown in Fcn.~\ref{Fcn:fit3}. Besides the filter number $\hat{N}_i$ of a layer, $\hat{N}_{i-1}$ also has a greater impact on $r_{s_i}$, suggesting that it is very difficult to directly find an optimal compact architecture of the original network. Moreover, excavating redundancy in the filter itself may be a more promising way to speed it up, \eg if we discard half of the filters per layer, the speed-up ratio of the proposed method is about $4\times$.

\textbf{Compression ratio.} The compression ratio on convolution filters is easy to calculate and is equal to the last term in Fcn.~\ref{Fcn:fit3}. Specifically, for the $i$-th convolutional layer, the compression ratio of the proposed method is
\begin{equation}
r_{c_i} = \frac{H_iW_iN_{i-1}N_i}{H_iW_i\hat{N}_{i-1}\hat{N}_i} = \frac{N_{i-1}N_i}{\hat{N}_{i-1}\hat{N}_i}.
\end{equation}
However, besides the convolution filters, there are other memory usages that are often ignored in existing methods. In fact, the feature maps of different layers account for a large proportion of online memory. In some implementations~\cite{MatConvnet}, the feature maps of a layer are removed after they have been used to calculate the following layer to reduce the online memory usage. However, memory allocation and release are time consuming. In addition, short-cut layers are widely used in recent CNN models~\cite{ResNet}, in which previous feature maps are preserved for combination with other layers. Discarding redundant convolutional filters significantly reduces the memory usage of feature maps. For a given convolutional layer $\mathcal{L}_i$, the compression ratio of the proposed method on feature maps is 
\begin{equation}
r_{f_i} = \frac{N_iH'_iW'_i}{\hat{N}_iH'_iW'_i} = \frac{N_i}{\hat{N_i}},
\label{Fcn:rf}
\end{equation} 
which is directly affected by the sparsity of $\mathbf{b}_i$. Accordingly, the memory to store the feature maps of other layers (\eg pooling and ReLU) will be reduced simultaneously. The experimental results and a discussion of compression and speed-up ratios are presented in the following section.

\section{Experiments}
\textbf{Baseline methods and Datasets.}
The proposed method was evaluated on four baseline CNN models: LeNet~\cite{LeNet}, AlexNet~\cite{AlexNet}, VGGNet-16~\cite{VGGNet}, and ResNet-50~\cite{ResNet}, and conducted using the MNIST handwritten digit and ILSVRC datasets. We used MatConvNet~\cite{MatConvnet} and NVIDIA Titan X graphics cards to implement the proposed method. In addition, several state-of-the-art approaches were selected for comparison including P+QH (Pruning + Quantization and Huffman encoding)~\cite{pruning}, SVD~\cite{SVD}, XNOR-Net~\cite{XNORNet}, and CNNpack~\cite{CNNpackNIPS}.

\textbf{LeNet on MNIST.} 
The performance of the proposed method was first evaluated on a small network to study some of its characteristics. The network has four convolutional layers of size $5\times5\times1\times20$, $5\times5\times20\times50$, $4\times4\times50\times500$, and $1\times1\times500\times10$, respectively. The model was trained with batch normalization layers and the accuracy was 99.20\%.

The proposed method has several parameters as shown in Alg.~\ref{Alg:main}. Population $K$ was set to 1000 to ensure a sufficiently large search space, and the maximum iteration number $T$ was set to 100. Three probability parameters were empirically set to $s_1 = 0.2$, $s_2 = 0.7$, and $s_3 = 0.1$~\cite{GA}. A larger $\lambda$ makes the compressed network more compact, but the accuracy of the original network cannot be retained in the compressed counterpart. We tuned this parameter from $0.5$ to $1.5$ and set it to $0.9$, which was the best trade-off between network accuracy and compression ratio, since the accuracy of the original network was very high and each individual could maintain considerable accuracy.

The evolutionary process for compressing the network is shown in Fig.~\ref{Fig:diagram}. Individuals in the population are updated with higher fitness individuals after each iteration. As a result, we obtained a 103\emph{KB} compressed network that consistutes of four convolutional layers: $5\times5\times1\times9$, $5\times5\times9\times17$, $4\times4\times17\times84$, and $1\times1\times84\times10$, respectively. The model accuracy after fine-tuning was 99.20\%, \ie there was no decrease in accuracy. Compression and speed-up ratios of the entire network were $r_c = 15.52\times$, $r_s = 5.76$, and $r_f = 2.42$. %The MATLAB demo code for verifying the compressed network can be found in the Supplementary Material.

Furthermore, for fair comparison, we directly initialized a network with the same architecture and tuned it on MNIST. Unfortunately, the accuracy of this network was only $98.5\%$, significantly lower than that of the original network since it cannot inherit pre-trained convolution filters of the original network. Moreover, the accuracy of a network randomly generated with a similar number of filters was about $92.7\%$, demonstrating that the proposed method provides an effective architecture for constructing a portable network and inherits useful information from the original network.

\begin{figure}[t]
\centering
\includegraphics[width=0.95\linewidth]{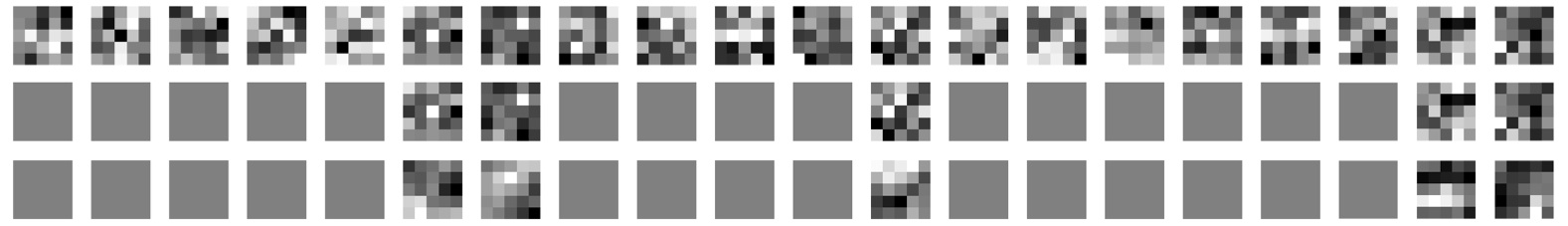}
\caption{Convolution filters learned on MNIST. From top to bottom: the original convolution filters, filters after applying the proposed method, and filters after fine-tuning.}
\label{Fig:filters}
\vspace{-1.5em}
\end{figure}

\textbf{Filter visualization.} 
The proposed method excavates redundant convolution filters using an evolutionary algorithm, which is different to the existing weight or filter pruning approaches. Therefore, it is necessary to explore which filters are recognized as redundant and which convolution filters are optimal for CNNs. To this end, we visualized the LeNet filters on MNIST before and after applying the proposed method, as shown in Fig.~\ref{Fig:filters}.

The result shown in the second row of Fig.~\ref{Fig:filters} is particularly interesting. Our method not only discards small filters but also removes some filters with large weights. Of note, the remaining filters after fine-tuning are even more distinct. The average Euclidean distance of filters in the third row is 0.2428, while the average cosine similarity of filters in the first and second rows are 0.1927 and 0.1789, respectively. This observation shows that redundancy can exist in either large or small convolution filters, and similar filters may be redundant and non-contributory to the entire CNN, providing a strong a priori rationale for designing and learning CNN architectures. In addition, the filters shown in the third line are obviously smoother than those in the original network, demonstrating the feasibility of compressing CNNs in the frequency domain as discussed in~\cite{CNNpackNIPS}.

\textbf{Compressing convolutional networks on ImageNet.}
We next employed the proposed evolutionary method (namely ECS) on ImageNet ILSVRC 2012~\cite{ImageNet}, which contains 1.2 millions images for training and 50k images for testing. We examined three mainstream CNN architectures: AlexNet~\cite{AlexNet}, VGGNet-16~\cite{VGGNet}, and ResNet-50~\cite{ResNet}. There are over 61M parameters in AlexNet and over 138M weights in VGGNet-16. ResNet-50 has about 25M parameters, which is more compact than the previous two CNNs. The top-5 accuracies of these three models were 80.8\%, 90.1\%, 92.9\%, respectively.

\begin{figure*}[t]
\centering
\includegraphics[width=1\linewidth]{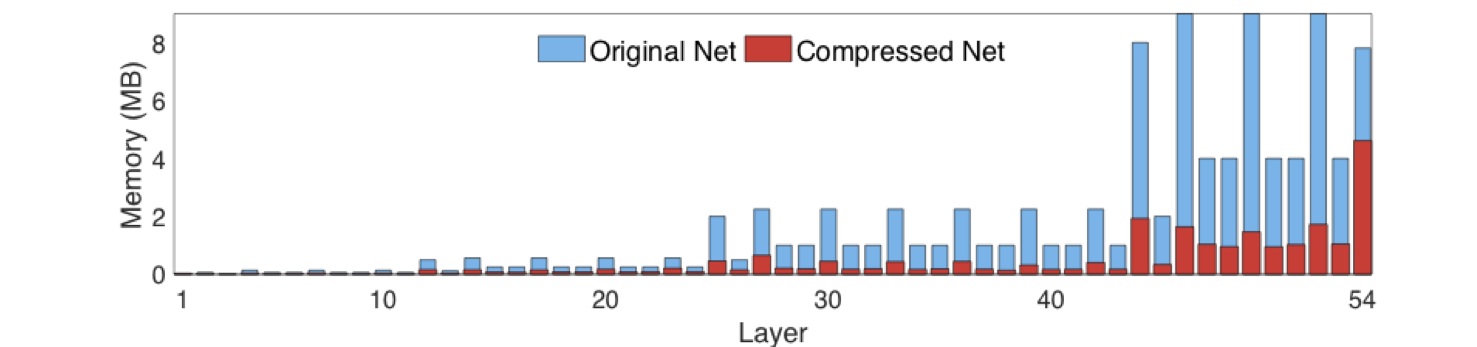}\\
(a) Compression ratio of all convolution filters ($r_c$).\\
\vspace{2em}
\includegraphics[width=1\linewidth]{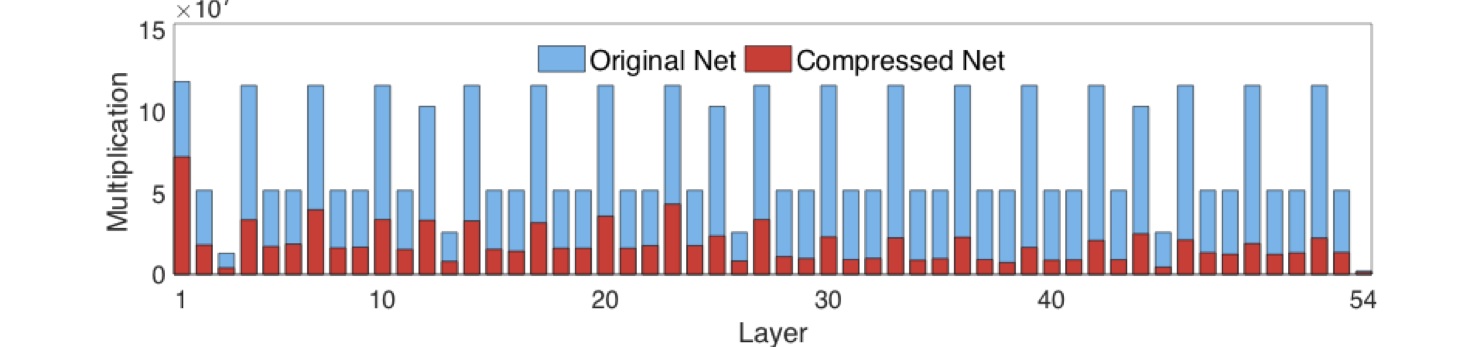}\\
(b) Speed-up ratio of all convolutional layers ($r_s$).
\caption{Compression statistics for ResNet-50.}
\label{Fig:Res}
\end{figure*}

Since the accuracy of convolutional networks on ImageNet was much harder to maintain, we adjusted $\lambda = 0.5$ to allow individuals higher accuracy evolution. The compression and speed-up ratios of the proposed methods on the three CNNs are shown in Table~\ref{Tab:state-of-the-art}. In addition, architectures of compressed AlexNet and VGGNet-16 are shown in Tab.~\ref{Tab:AlexNet} and Tab.~\ref{Tab:VGG}, and detailed compression and speed-up results of ResNet-50 are shown in Fig.~\ref{Fig:Res}.

\begin{table}[t]
\centering
\caption{An overall comparison between state-of-the-art CNN compression methods and the proposed evolutionary compression scheme (ECS) on the ILSVRC2012 dataset. The overall compression and speed-up ratios are denoted $r_c$ and $r_s$, respectively. }
\label{Tab:state-of-the-art}
\vspace{0.5em}
\scriptsize
\begin{tabular}{c||c|c|c|c|c|c|c}
\hline
Model & Eval. & Orig. & P+QH~\cite{pruning} & SVD~\cite{SVD} & \footnotesize{Perfor.}~\cite{sparse1}& \footnotesize{CNNpack}~\cite{CNNpackNIPS}& ECS\\
\hline \hline
\multirow{4}{1.35cm}{AlexNet~\cite{AlexNet}} & $r_c$ & 1 &  35$\times$ & 5$\times$ & 1.7$\times$ & 39$\times$& 5.00$\times$\\
&$r_s$ & 1 & - & 2$\times$ & 2$\times$& 25$\times$& 3.34$\times$\\
& $top$-$1\ err$ & 41.8\%  & 42.7\% & 44.0\% & 44.7\% & 41.6\%& 41.9\%\\
& $top$-$5\ err$ & 19.2\%& 19.7\% & 20.5\% & - & 19.2\%& 19.2\% \\
\hline
\multirow{4}{1.35cm}{VGGNet~\cite{VGGNet}} & $r_c$ & 1 & 49$\times$ & - & 1.7$\times$ & 46$\times$& 8.81$\times$\\
&$r_s$ & 1 &  3.5$\times$& - & 1.9$\times$ & 9.4$\times$& 5.88$\times$\\
& $top$-$1\ err$ & 28.5\% & 31.1\% & - & 31.0\%& 29.7\% &29.5\%\\
& $top$-$5\ err$ & 9.9\% & 10.9\% & - & - & 10.4\%& 10.2\%\\
\hline
\multirow{4}{1.35cm}{ResNet~\cite{ResNet}} & $r_c$ & 1 &- & - & - & 12.2$\times$ &4.10$\times$\\
&$r_s$ & 1 &  -& - & - & 4$\times$& 3.83$\times$\\
& $top$-$1\ err$ & 24.6\% &  - & - & - & -& 25.1\%\\
& $top$-$5\ err$ & 7.7\% &  - & - & - & 7.8\%&8.0\%\\
\hline
\end{tabular}
%\vspace{-1.5em}
\end{table}

\begin{table}[h]
\centering
\caption{Compression statistics for AlexNet.}
\label{Tab:AlexNet}
\scriptsize
\vspace{0.5em}
\begin{tabular}{c||c|c|c|c|c}
\hline
Layer & Num of Weights & Memory & Num of New Weights&Memory &$r_{c}$ \\
\hline \hline
conv1& $11\times11\times3\times96$&1.24MB &$11\times11\times3\times56$ & 0.08MB& $1.71\times$ \\
conv2& $5\times5\times48\times256$&1.88MB &$5\times5\times28\times120$ &  0.32MB &$3.66\times$ \\
conv3& $3\times3\times256\times384$&3.62MB &$3\times3\times120\times190$ &  0.78MB & $4.31\times$  \\
conv4& $3\times3\times192\times384$&2.78MB &$3\times3\times95\times188$ &  0.61MB & $4.12\times$  \\
conv5& $3\times3\times94\times144$&1.85MB&$3\times3\times175\times226$ &  0.46MB & $3.63\times$  \\
fc6& $6\times6\times256\times4096$& 144MB&$6\times6\times144\times1386$  & 27.41MB & 5.25$\times$  \\
fc7& $1\times1\times4096\times4096$&64MB &$1\times1\times1386\times1848$ &  9.77MB & 6.55$\times$  \\
fc8& $1\times1\times4096\times1000$&15.62MB &$1\times1\times1848\times1000$ &  7.05MB & 2.22$\times$ \\
\hline
Total& 60954656 & 232.52MB & 12186444 & 46.48MB & 5.00$\times$ \\
\hline
\end{tabular}
\vspace{1em}
\caption{Compression statistics for VGG-16 Net.}
\label{Tab:VGG}
\vspace{0.5em}
\begin{tabular}{c||c|c|c|c|c}
\hline
Layer & Num of Weights & Memory & Num of New Weights&Memory &$r_c$ \\
\hline \hline
conv1\_1 & $3\times3\times3\times64$&12.26MB &$3\times3\times3\times12$ & 0.001MB& 5.33$\times$  \\
conv1\_2 & $3\times3\times64\times64$&12.39MB &$3\times3\times12\times28$ & 0.01MB&12.19$\times$  \\
conv2\_1& $3\times3\times64\times128$&6.41MB &$3\times3\times28\times57$ & 0.05MB&5.13$\times$  \\
conv2\_2 & $3\times3\times128\times128$&6.69MB &$3\times3\times57\times61$ & 0.12MB&4.71$\times$  \\
conv3\_1& $3\times3\times128\times256$&4.19MB &$3\times3\times61\times133$ & 0.28MB& 4.04$\times$  \\
conv3\_2& $3\times3\times256\times256$&5.31MB &$3\times3\times133\times127$ & 0.58MB& 3.88$\times$  \\
conv3\_3&$3\times3\times256\times512$&5.31MB &$3\times3\times127\times137$ & 0.60MB& 3.77$\times$  \\
conv4\_1& $3\times3\times512\times512$&6.03MB &$3\times3\times137\times194$ & 0.91MB& 4.93$\times$   \\
conv4\_2& $3\times3\times512\times512$&10.53MB &$3\times3\times194\times119$ & 0.79MB& 11.36$\times$   \\
conv4\_3& $3\times3\times512\times512$&10.53MB &$3\times3\times119\times320$ & 1.31MB& 6.88$\times$  \\ 
conv5\_1& $3\times3\times512\times512$&9.38MB &$3\times3\times320\times72$ & 0.79MB& 11.38$\times$   \\
conv5\_2& $3\times3\times512\times512$&9.38MB &$3\times3\times72\times62$ & 0.15MB& 58.72$\times$   \\
conv5\_3&$3\times3\times512\times512$&9.38MB &$3\times3\times62\times122$ & 0.26MB& 34.66$\times$   \\
fc6 & $7\times7\times512\times4096$&392MB &$7\times7\times122\times2300$ & 52.45MB& 7.47$\times$ \\
fc7 & $1\times1\times4096\times4096$&64MB &$1\times1\times2300\times125$ & 1.09MB& 58.36$\times$   \\
fc8 & $1\times1\times4096\times1000$&15.62MB &$1\times1\times125\times1000$ & 0.48MB& 32.77$\times$  \\
\hline
Total & $138344128$&579.46MB& $20118610$ & 59.88MB& 8.81$\times$  \\ 
\hline
\end{tabular}
\end{table}

A detailed comparison of the above three benchmark deep CNN models between the proposed method and state-of-the-art methods can also be found in Table~\ref{Tab:state-of-the-art}. Since the convolution filters learned by ECS in the compressed networks are significantly reduced and the number of channels per layer is simultaneously decreased, we obtained higher speed-up ratios on these CNNs according to Fcn.~\ref{Fcn:rs}. The compression ratios of the proposed method were lower than those of the other approaches, because the weights of the compressed networks are still stored in 32-bit floating numbers, which can be directly implemented in most existing devices. For fair comparison, we used 8-bit floating numbers to quantize followed by a fine-tuning process to further enhance compression performance, suggested by~\cite{CNNpackNIPS}. The results are encouraging, with the compressed networks with 8-bit floating numbers having the same performance as their 32-bit versions. Therefore, the compression ratios of the proposed method should be multiplied by at least a factor of $4\times$, \eg we can obtain an about 16$\times$ compression ratio on the ResNet-50. This evaluation can be further increased by exploiting sparse shrinkage and Huffman encoding; however, these strategies have no contribution to online inference so we did not apply them here. When considering online inferences of deep CNNs, the proposed method is clearly the best approach for compressing convolutional networks. 

\textbf{Compression ratios on feature maps.} 
As discussed in Fcn.~\ref{Fcn:rf}, the compression ratio on CNN feature maps is also an important metric for evaluating compression methods, but it is ignored in most existing approaches. Therefore, the compression ratios on feature maps of these methods are both equal to 1$\times$ such as~\cite{SVD, CNNpackNIPS, pruning}.

\begin{table}[h]
\centering
\vspace{0.5em}
\caption{Compression ratios of feature maps on different CNN models.}
\small
\label{Tab:rf}
%\vspace{-0.5em}
%\small
\begin{tabular}{c||c|c|c|c}
\hline
Model & LeNet & AlexNet & VGGNet-16 & ResNet-50\\
\hline
\hline
Original memory & 0.07 \emph{MB} & 5.49 \emph{MB}& 109.26 \emph{MB}& 137.25 \emph{MB}\\
\hline
Compression ratio $r_f$ & 2.42$\times$ & 1.88$\times$& 2.54$\times$& 1.86$\times$ \\
\hline
\end{tabular}
%\vspace{-0.5em}
\end{table}

The compression ratios on feature maps of the proposed method on different CNNs are shown in Table~\ref{Tab:rf}. It is clear that the compressed networks are more portable and can be directly used for online inference without any additional technical support since they are still regular CNN models.

\section{Discussion and Conclusions}
CNNs with higher performance and portable architectures are urgently required for mobile devices. This paper presents an effective CNN compression technique using an evolutionary algorithm. Compared to state-of-the-art methods, we no longer directly recognize some weights or filters as redundant. The proposed method identifies redundant convolution filters by iteratively refining a certain number of networks before learning a compressed network with significantly fewer parameters. Our experiments show that the proposed method can achieve significant compression and speed-up ratios and retain the classification accuracy of the original neural network. Moreover, the network compressed by the proposed approach is still a regular CNN that can be directly used for online inference without any decoding.
\clearpage
{\small
\bibliographystyle{plain}
\bibliography{ref}}

\end{document}